\documentclass[10pt,journal,compsoc]{IEEEtran}
\ifCLASSOPTIONcompsoc
  \usepackage[nocompress]{cite}
\else
  \usepackage{cite}
\fi

\usepackage{textcomp}
\usepackage{multirow}
\usepackage{graphicx}

\usepackage[numbers,sort]{natbib}

\usepackage[normalem]{ulem}
\usepackage{subfigure}
\usepackage{url}
\usepackage{booktabs}
\usepackage{cite}
\usepackage{ifsym}
\usepackage{comment}
\usepackage{times}
\usepackage{soul}
\usepackage[hidelinks]{hyperref}
\hypersetup{
    colorlinks=true,
    linkcolor=red,
    citecolor=green,
}
\usepackage[utf8]{inputenc}
\usepackage[small]{caption}
\usepackage{amsmath}
\usepackage{amsthm}
\usepackage{algorithm}
\usepackage{algorithmic}
\usepackage[switch]{lineno}
\usepackage{tikz}
\usepackage[edges]{forest}
\usepackage{subfigure}
\usepackage{cleveref}
\usepackage{enumitem} % leftmargin

\renewcommand{\IEEEbibitemsep}{0pt plus 0.5pt}
\makeatletter
\IEEEtriggercmd{\reset@font\normalfont\fontsize{6.55pt}{6.55pt}\selectfont} % !!!! need to be removed later
\makeatother
\IEEEtriggeratref{1}

\begin{document}
\title{Graph World Models: Concepts, Taxonomy, and Future Directions}

 \author{Jiawei~Liu,~Senqiao~Yang,~Mingjun~Wang,~Yu~Wang,~Bei~Yu 
 \IEEEcompsocitemizethanks{
 \IEEEcompsocthanksitem Jiawei~Liu,~Senqiao~Yang,~Mingjun~Wang,~Bei~Yu are with The Chinese University of Hong Kong, Hong Kong, China. 
%\protect\\
% % note need leading \protect in front of \\ to get a newline within \thanks as
% % \\ is fragile and will error, could use \hfil\break instead.
% E-mail: liujw@cse.cuhk.edu.hk

% %\IEEEcompsocthanksitem J. Liu is with Singapore University of Technology and Design, Singapore.% <-this % stops an unwanted space
%\protect\\
% %E-mail: jun\_liu@sutd.edu.sg.

\IEEEcompsocthanksitem Yu Wang is with Tsinghua University, Beijing, China. 

}

}

\IEEEtitleabstractindextext{%
\begin{abstract}
As one of the mainstream models of artificial intelligence, world models allow agents to learn the representation of the environment for efficient prediction and planning. However, classical world models based on flat tensors face several key problems, including noise sensitivity, error accumulation and weak reasoning.
To address these limitations, many recent studies use graph structure to decompose the environment into entity nodes and interactive edges, and model virtual environments in a structured space. This paper systematically formalizes and unifies these emerging graph-based works under the concept of graph world models (GWMs). To the best of our knowledge, GWMs have not yet been explicitly defined and surveyed as a unified research paradigm. Furthermore, we propose a taxonomy based on relational inductive biases (RIB), categorizing GWMs by the specific structural priors they inject: (1) spatial RIB for topological abstraction; (2) physical RIB for dynamic simulation; and (3) logical RIB for causal and semantic reasoning. For each model category, we outline the key design principles, summarize representative models, and conduct comparative analyses. We further discuss open challenges and future directions, including dynamic graph adaptation, probabilistic relational dynamics, multi-granularity inductive biases, and the need for dedicated benchmarks and evaluation metrics for GWMs.
\end{abstract}

}

\maketitle

\IEEEdisplaynontitleabstractindextext

% \fi
\IEEEpeerreviewmaketitle

\renewcommand{\baselinestretch}{0.9} % !!!! need to be removed later !!!!!!!!!!

\section{Introduction}

As an important branch of artificial intelligence, world models learn the compressed spatio-temporal representation of the environment~\cite{ha2018world,ha2018recurrent}. Based on its vision module for perception and memory module for dynamic prediction, agents can simulate future states and learn decisions within world models rather than directly interacting with the real world. Therefore, world models show great application potential in the fields of robotics~\cite{hafner2025mastering}, autonomous driving~\cite{guan2024world}, and video generation~\cite{OpenAI2024Sora}.

However, there are several limitations of the early world model architecture that hinder deployment in complex and high-risk environments. (1) \textit{Noise sensitivity}: Pixel-level modeling consumes a lot of capacity to model non-task-related details such as background noise, making it difficult for the model to capture critical structured features~\cite{savinov2018sptm,zhang2021l3p,taniguchi2021point}. (2) \textit{Error accumulation}: In long-range virtual simulation, the prediction deviation of each step will continue to accumulate over time, causing the prediction trajectory to deviate rapidly from the real environment. \cite{kipf2020cswm,li2020vgpl,nanbo2025facts}. (3) \textit{Weak reasoning}: Due to the lack of modeling of object interaction laws, it is difficult for the model to perform effective logical deduction and cross-task generalization in complex interaction scenarios. \cite{wu2024coke,feng2025gwm}.

\begin{figure}[t]
\centering
\includegraphics[width=\linewidth]{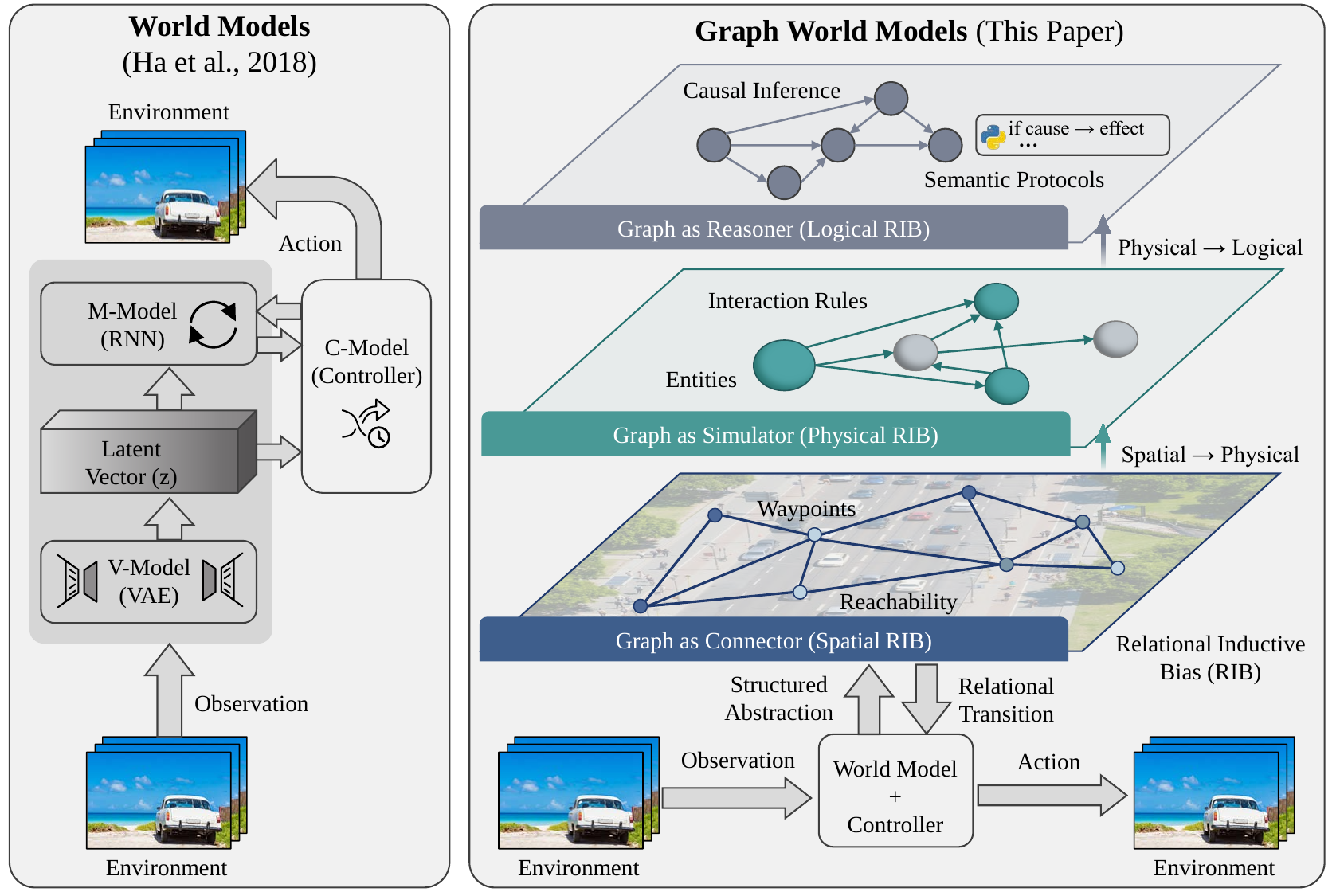}
\caption{Comparison of the classical world models (WMs) and the graph world models (GWMs). (Left) The classical world model represents the environment through the unstructured latent vector. (Right) GWMs extend WMs by injecting relational inductive biases from spatial, physical, or logical aspects.}
\label{fig: intro}
\end{figure}

To address these limitations, recent work has increasingly employed structured state representations to enhance world models. However, despite the emergence of this paradigm, the community still lacks a common definition. In this paper, we refer to these works as graph world models (GWMs), and formalize GWMs by structurally injecting relational inductive bias (RIB)~\cite{battaglia2018relational} into the input environment of the world model. As shown by~\Cref{fig: intro}, GWM models the environment as a graph $\mathcal{G} = (\mathcal{V}, \mathcal{E})$, and the latent relationship is modelled by edges. In this way, GWM can inject relational inductive biases from spatial, physical, or logical aspects, mitigating the above-mentioned critical issues in classical world models.

Despite these emerging efforts, existing studies remain scattered across reinforcement learning, robotics, computer vision, embodied AI, and LLM-based agents. Existing surveys on world models, robotic world models, video generation, and embodied agents provide valuable overviews, but they do not explicitly formalize graph world models as a unified research paradigm centered on graph-structured relational inductive biases. Unlike the general survey's taxonomy of world models \cite{sakagami2023robotic,guan2024world,cho2024sora,ding2025understanding,fung2025embodied,chu2026awm}, we propose a three-layer taxonomy that organizes GWMs by the RIBs they inject—spatial, physical, or logical. This taxonomy not only demonstrates the role of graphs in world models, but is also relevant to model capabilities and their concerns:

\begin{itemize}[leftmargin=*, itemsep=-1pt]
\item Graph as Connector (Spatial RIB): uses graphs to model reachability and connectivity.
\item Graph as Simulator (Physical RIB): uses graphs to distill complex physics into interaction rules.
\item Graph as Reasoner (Logical RIB): uses graphs to extract semantic protocols or causal skeletons.
\end{itemize}

Based on this definition and taxonomy, this survey makes three contributions:
\begin{enumerate}[leftmargin=*, itemsep=-1pt]
\item To the best of our knowledge, this is the first survey to explicitly formalize graph world models as a unified research paradigm. Our definition provides a mathematical framework that unifies diverse research lines.
\item This paper introduces a novel taxonomy that categorizes GWMs into connector, simulator, and reasoner based on RIBs. Based on this taxonomy, we conduct thorough research and comprehensive discussions on relevant papers published in top conferences and journals.
\item This article provides promising future directions to advance the field of GWMs.
\end{enumerate}
\section{Definitions and Taxonomy}

\subsection{Formal Definitions}
As proposed by the original paper \cite{ha2018world,ha2018recurrent}, a world model (WM) is a model that can be trained quickly in an unsupervised manner to learn a compressed spatial and temporal representation of the environment, which can be formally defined as follows.

\textbf{Definition 1 (World Model)}. The world model is a triplet $\mathcal{M} = \langle \mathcal{S}, V, M \rangle$, where $\mathcal{S} = \mathcal{Z} \times \mathcal{H}$ is the latent state space. It operates on observation $o_t \in \mathcal{O}$ and action $a_t \in \mathcal{A}$ for each timestep $t$ with two modules:

\begin{itemize}[leftmargin=*, itemsep=-1pt]
\item \textbf{Vision Module ($V$)}: $z_t = V(o_t)$, mapping the original observation data into a latent feature $z_t \in \mathcal{Z}$, essentially extracting key information from the original data.
  \item \textbf{Memory Module ($M$)}: $h_{t+1} = M(z_t, a_t, h_t)$, where the hidden state $h_t \in \mathcal{H}$ aggregates historical information. Based on the updated hidden state $h_{t+1}$, the model predicts the next latent feature $\hat{z}_{t+1} \sim P(z_{t+1}|h_{t+1})$ through the probability distribution $P(\cdot)$, which describes the state transition law of the environment.

\end{itemize}

At any time step $t$, $(z_t, h_t)$ together constitute the complete state of the internal world. The external controller $C$ will learn the decision-making strategy $\pi(a_t|s_t)$ by interacting with the simulation scenario generated by $M$, interacting with the environment as an output interface for the agent.

\textbf{Definition 2 (Graph World Model)}. The graph world model (GWM) is an extension of the classic world model. Its core is to represent the environment as a graph structure $\mathcal{G}_t = (\mathcal{V}_t, \mathcal{E}_t) \in \mathcal{G}$, where $\mathcal{V}_t$ is a node set and $\mathcal{E}_t$ is an edge set. Correspondingly, it includes two core operations:

\begin{itemize}[leftmargin=*, itemsep=-1pt]
  \item \textbf{Structural Abstraction ($\psi$)}: $\mathcal{G}_t = \psi(o_t)$, constructing graphs from the original observation data $o_t$ or latent state $s_t$ at a specific initial time $t=t_{init}$.

  \item \textbf{Relational Transition ($\mathcal{T}_G$)}: $\mathcal{G}_{t+1} = \mathcal{T}_G(\mathcal{G}_t, a_t, h_t)$, modelling the dynamic change process of the graph structure and attributes at subsequent time $t\geq t_{init}$.

\end{itemize}

\subsection{Taxonomy}

\begin{figure*}[t!]
\centering
\resizebox{\textwidth}{!}{
\tikzset{
    basic/.style  = {draw=black, text width=2cm, align=center, rectangle, thin, rounded corners=2pt},
    % 黑白灰色系：一级节点（主标题，深灰）
    root/.style   = {basic, fill=gray!30, text width=0.5cm,},
    % 黑白灰色系：二级节点，中灰
    node1/.style = {basic, fill=gray!20, text width=2cm,},
    % 黑白灰色系：三级节点，浅灰（保持5cm加宽宽度）
    node11/.style = {basic, fill=gray!15, text width=5cm,},
    node12/.style = {basic, fill=gray!15, text width=5cm,},
    node13/.style = {basic, fill=gray!15, text width=5cm,},
    node14/.style = {basic, fill=gray!15, text width=5cm,},
    node15/.style = {basic, fill=gray!15, text width=5cm,},
    node16/.style = {basic, fill=gray!15, text width=5cm,},
    % 黑白灰色系：叶子节点，极浅灰（保持13cm宽度适配页面）
    node111/.style = {basic, fill=gray!10, text width=13cm,},
    node121/.style = {basic, fill=gray!10, text width=13cm,},
    node131/.style = {basic, fill=gray!10, text width=13cm,},
    node141/.style = {basic, fill=gray!10, text width=13cm,},
    node151/.style = {basic, fill=gray!10, text width=13cm,},
    node161/.style = {basic, fill=gray!10, text width=13cm,},
    % 全局边样式：一级-二级-三级直角分叉
    edge from parent/.style={draw=black, edge from parent fork right}
}
\begin{forest} for tree={
    grow=east,
    growth parent anchor=west,
    parent anchor=east,
    child anchor=west,
    % 全局边路径：一级-二级-三级直角分叉
    edge path={\noexpand\path[\forestoption{edge}] (!u.parent anchor) -- +(10pt, 0) |- (.child anchor) \forestoption{edge label};},
    l sep=7mm,
    calign=center,  % 强制父子节点垂直居中，为纯水平连线提供保障
    s sep=3mm,
}
% 一级节点：Graph World Model
[\rotatebox{90}{Graph World Model}, root, l sep=7mm, yshift = -16.3mm
    % 二级节点1：Graph as Reasoner
    [Graph as Reasoner, node1, l sep=7mm
        % 三级节点2：Latent Causal Identification（对应叶子节点：2行）
        [Invariant Causal Skeletons, node15, l sep=5mm, yshift = -0.3mm
            % 叶子节点2：2行（从上到下第2个，手动拆分2行适配页面）
            [{FANS-RL~\cite{feng2022fansrl}, VCD~\cite{lei2023vcd}, FACTS \cite{nanbo2025facts},  CCSA~\cite{zhao2025ccsa}}, node151,
             % 关键修改：覆盖分叉样式+纯水平路径
             edge from parent/.style={draw=black},
             edge path={\noexpand\path[\forestoption{edge}] (!u.parent anchor) -- (.child anchor) \forestoption{edge label};}]
        ]
        % 三级节点1：Symbolic and Social Protocols（对应叶子节点：3行）
        [Normative Semantic Protocols, node16, l sep=5mm, yshift = -4mm
            % 叶子节点1：3行（从上到下第1个，手动拆分3行适配页面）
            [{Worldformer~\cite{ammanabrolu2021worldformer}, S3~\cite{gao2023s3}, COKE~\cite{wu2024coke}, FPWC~\cite{yin2025fpwc}, DAVIS~\cite{dinh2025davis}, AriGraph~\cite{anokhin2025arigraph}, OpenFunGraph~\cite{zhang2025openfungraph}, SWMPO \cite {cano2025swmpo},
            \\GWM~\cite{feng2025gwm}, Polanyi~\cite{de2025polanyi}, YuLan-OneSim~\cite{wang2025yulan}}, node161,
             % 关键修改1：覆盖全局分叉样式，取消分叉逻辑
             edge from parent/.style={draw=black},
             % 关键修改2：纯水平直线路径，直接连接父子锚点，无弯折
             edge path={\noexpand\path[\forestoption{edge}] (!u.parent anchor) -- (.child anchor) \forestoption{edge label};}]
        ]
    ]
    % 二级节点2：Graph as Simulator
    [Graph as Simulator, node1, l sep=7mm
        [System-centric Interaction, node14, l sep=5mm, yshift = 0.1mm
            % 叶子节点3：2行（从上到下第3个，手动拆分2行适配页面）
            [{VDFD~\cite{wang2023vdfd}, RoboPack~\cite{ai2024robopack}, HD-VPD~\cite{whitney2025hdvpd}}, node141,
             % 关键修改：覆盖分叉样式+纯水平路径
             edge from parent/.style={draw=black},
             edge path={\noexpand\path[\forestoption{edge}] (!u.parent anchor) -- (.child anchor) \forestoption{edge label};}]
        ]
        [Object-centric Interaction, node13, l sep=5mm, yshift = -1.9mm
            % 叶子节点4：2行（从上到下第4个，手动拆分2行适配页面）
            [{G-SWM~\cite{lin2020gswm}, C-SWM~\cite{kipf2020cswm}, CWM~\cite{li2020cwm}, VGPL~\cite{li2020vgpl},  GATSBI~\cite{min2021gatsbi}, 3D-OES~\cite{tung20213does}, CEE-US~\cite{sancaktar2023ceeus}, ROCA~\cite{ugadiarov2025roca}, FACTS~\cite{nanbo2025facts}, FIOC-WM~\cite{feng2025fiocwm}, Dyn-O~\cite{wang2025dyno}}, node131,
             % 关键修改：覆盖分叉样式+纯水平路径
             edge from parent/.style={draw=black},
             edge path={\noexpand\path[\forestoption{edge}] (!u.parent anchor) -- (.child anchor) \forestoption{edge label};}]
        ]
    ]
    % 二级节点3：Graph as Connector
    [Graph as Connector, node1, l sep=7mm
        % 三级节点1：Implicit Experiential Memory（对应叶子节点：2行）
        [Implicit Experiential Memory, node12, l sep=5mm, yshift = 0.1mm
            % 叶子节点5：2行（从上到下第5个，手动拆分2行适配页面）
            [{WGD~\cite{shang2019worlddiscovery}, CSTIG~\cite{verbelen2022cstig}, VMG~\cite{zhu2023vmg}, GBMR~\cite{kang2023gbmr}, RLF~\cite{hu2024rlf}, G4RL~\cite{zhang2025g4rl}}, node121,
             % 关键修改：覆盖分叉样式+纯水平路径
             edge from parent/.style={draw=black},
             edge path={\noexpand\path[\forestoption{edge}] (!u.parent anchor) -- (.child anchor) \forestoption{edge label};}]
        ]
        % 三级节点2：Explicit Spatial Topology（对应叶子节点：2行）
        [Explicit Spatial Topology, node11, l sep=5mm, yshift = -3.7mm
            % 叶子节点6：2行（从上到下第6个，手动拆分2行适配页面）
            [{SPTM~\cite{savinov2018sptm}, SoRB~\cite{eysenbach2019sorb}, POINT~\cite{taniguchi2021point}, 
            L3P~\cite{zhang2021l3p}, PPGS~\cite{bagatella2022ppgs},  RGL~\cite{bonnavaud2023rgl},  \\Dreamwalker~\cite{wang2023dreamwalker}, 
            OVER-NAV~\cite{zhao2024overnav}, 
            AnyHome~\cite{fu2024anyhome}, ConceptGraphs~\cite{gu2024conceptgraphs},
            \\Open3DSG~\cite{koch2024open3dsg}, CityNavAgent~\cite{zhang2025citynavagent},
            OSU-3DSG~\cite{yu2025osu3dsg}
            }, node111,
             % 关键修改：覆盖分叉样式+纯水平路径
             edge from parent/.style={draw=black},
             edge path={\noexpand\path[\forestoption{edge}] (!u.parent anchor) -- (.child anchor) \forestoption{edge label};}]
        ]
    ]
]
\end{forest}
}
\caption{Taxonomy and representative papers on GWMs. Each work is placed according to its dominant relational inductive bias, while some recent models may naturally span multiple categories.}
\label{fig: taxonomy}
\end{figure*}
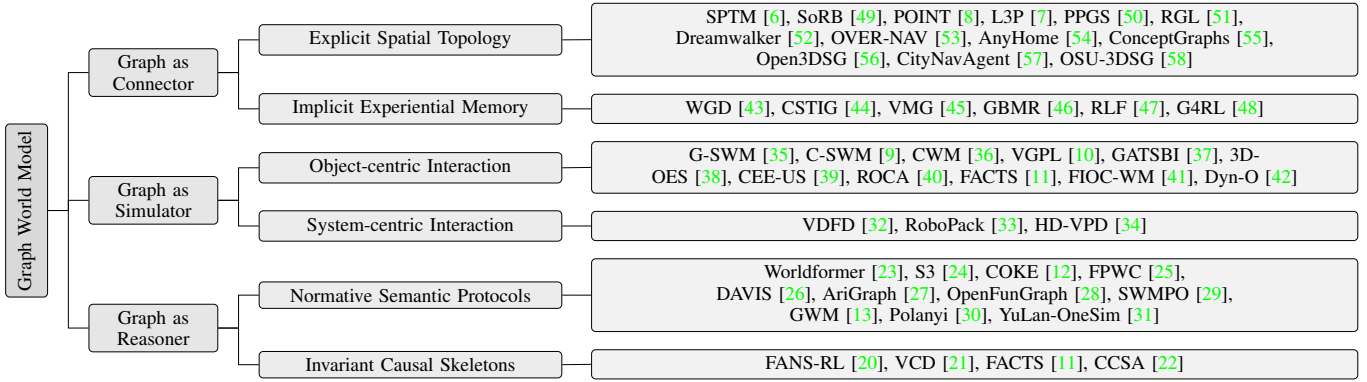

In GWMs, environment-to-graph structural abstraction can be seen as the process of injecting relational inductive biases into the environment. Different kinds of relationships give the model different capabilities. In this paper, we divide GWMs into three levels from the perspective of RIB, corresponding to three categories, as shown in \Cref{fig: taxonomy}.

\noindent\textbf{First layer: graph as connector.} Based on spatial RIB, these models focus on describing the reachability relationships and compressing high-dimensional trajectory data.

\noindent\textbf{Second layer: graph as simulator.} Based on physical RIB, these models focus on analyzing the dynamic laws of the environment and refining physical rules into state transitions.

\noindent\textbf{Third layer: graph as reasoner.} Based on logical RIB, these models focus on abstracting the semantic rules or causality, supporting instruction following and reasoning.

\noindent\textbf{Remark on category boundaries.} The proposed taxonomy is not intended to be strictly mutually exclusive. Instead, it characterizes GWMs according to their dominant relational inductive bias. In practice, some recent models may span multiple categories. For example, a model may use object-centric physical interactions for simulation while also maintaining causal factors for reasoning, or combine spatial scene graphs with semantic protocols for long-horizon planning. Such overlaps do not weaken the taxonomy; rather, they reveal an emerging trend toward hybrid GWMs that jointly integrate spatial, physical, and logical relational structures.
\section{Graph as Connector}

\label{sec: connector}

\begin{figure}[t]

\centering

\includegraphics[width=\linewidth]{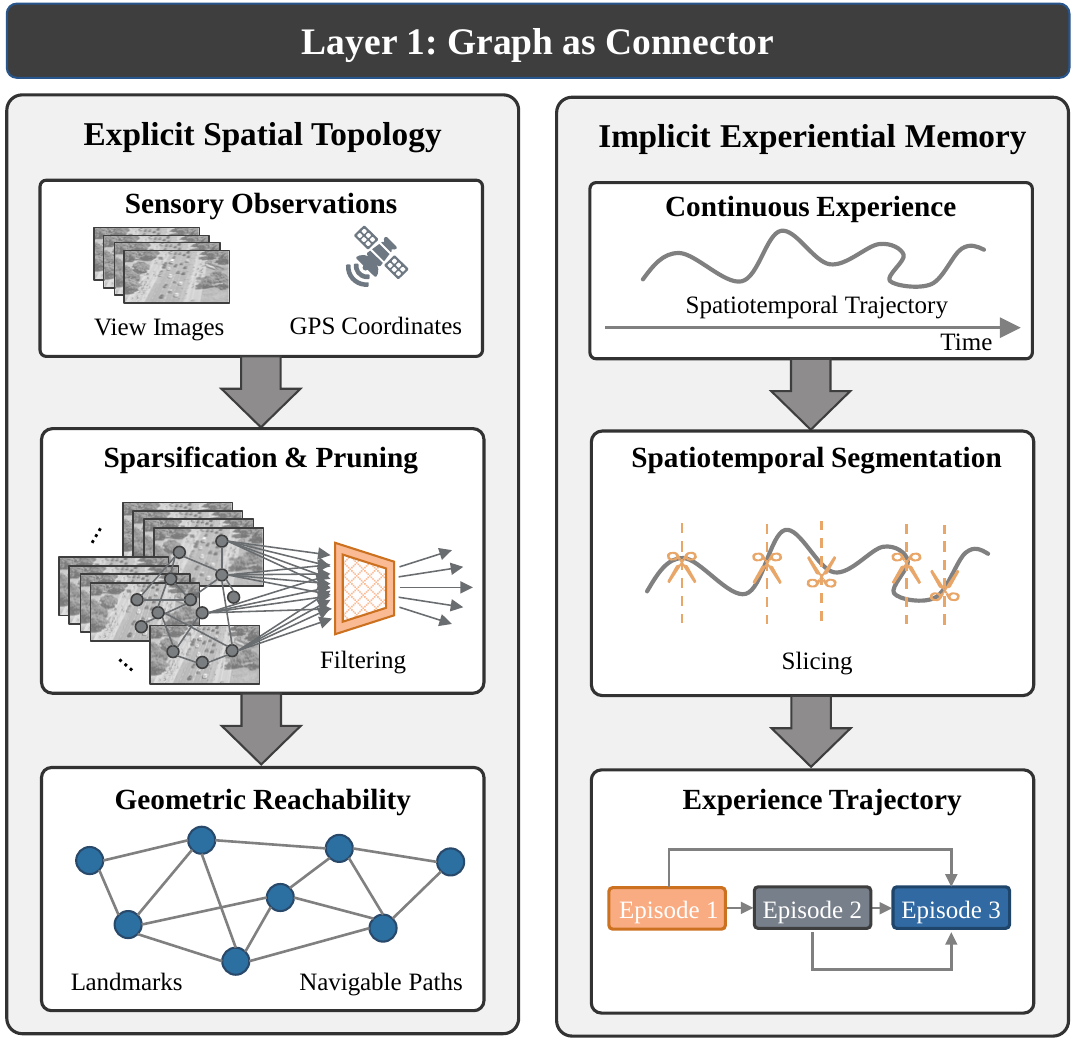}

\caption{Illustration of graph as connector. (Left) \textit{Explicit Spatial Topology}. (Right) \textit{Implicit Experiential Memory}.}

\label{fig: connector}

\end{figure}

In the continuous space with pixels as the core, there is not only a large amount of noise, but also a combinative increase in search complexity~\cite{ha2018world,shang2019worlddiscovery}. At the connector level, GWMs construct a spatial topology framework, which mainly solves the problems of accessibility in noise-sensitive and long-term scenarios. Specifically, these methods abstract the sensing data stream or experience fragment into nodes representing waypoints or landmarks and edges representing positional relationships. As shown in \Cref{fig: connector}, connector methods can be divided into two types of paradigms: (i) \textit{explicit spatial topology} (see~\Cref{sec: connector-explicit}), which maps observations directly to landmarks and their connections; (ii) \textit{implicit experimental memory} (see~\Cref{sec: connector-implicit}), which cuts trajectories into semantic fragments representing memory units.

\subsection{Explicit Spatial Topology}

\label{sec: connector-explicit}

Explicit spatial topology approaches are typically based on a predefined or clearly visible topology in the environment. Early topological-memory models such as SPTM~\cite{savinov2018sptm} provide one of the earliest examples of using graph-like structures as spatial substrates for planning, and can be viewed as foundational instances of connector-style GWMs. The SPTM model~\cite{savinov2018sptm} employs a semi-parametric topological structure consisting of two parts: a location graph with no static parameters and a learnable retrieval network. By calculating the similarity between the new observations and the data in the graph, the agent can plan the path based on the existing routing points in an unfamiliar environment.
Further, the SoRB method~\cite{eysenbach2019sorb} completely replaces the replay buffer in reinforcement learning (RL) with a graph structure. Each edge of the graph contains its weight, representing the policy difference between either state. However, once the replay-buffer coverage is low or the calculation difference is inaccurate, it will lead to some edges missing or wrong weights, which will lead to path search failure or poor path. To improve the robustness of the model, the POINT model~\cite{taniguchi2021point} uses omnidirectional observations and spherical convolution that is not affected by rotation to determine whether the target position can be reached from the current position, incorporating data augmentation and contrastive learning strategies.

To improve search efficiency while ensuring graph coverage, many subsequent studies have explored the automatic construction method of graph structure. For example, the L3P model \cite{zhang2021l3p} clusters the hidden space data of the model, identifies implicit landmarks representing key locations, and then eliminates redundant nodes. It should be noted that such landmarks are neither directly observable geometric features in images nor semantic information extracted from empirical fragments, but key anchors constrained by reachability. In this way, L3P can obtain a sparse set of landmarks, thereby narrowing the scope of subsequent graph searches. In addition, the PPGS method \cite{bagatella2022ppgs} constructs a state transition graph based on the extended scheme of autoregressive forward dynamics, and can also identify the repeated states that occur in the search process and cut off the loop, effectively improving the efficiency and flexibility of the planning process. In the face of the ever-changing environment, the RGL model \cite{bonnavaud2023rgl} dynamically adds and deletes nodes in the graph, ensuring that the reachable relationships within the graph are always accurate. For long-distance planning tasks, the Dreamwalker model \cite{wang2023dreamwalker} organizes the observation data into an environment graph and simulates unobserved scenes with the help of a scene generator to assist in completing the planning process based on Monte Carlo tree search. In order to enable the model to understand visual information and language instructions at the same time, the OVER-NAV method \cite{zhao2024overnav} integrates subgraphs from the agent's perspective into the navigation strategy, realizing iterative navigation that integrates visual perception and language understanding. The AnyHome model \cite{fu2024anyhome} uses large language models (LLMs) to generate bubble diagrams of graph structures, which are used as spatial constraints for scene generation tasks. For city-scale environments, the CityNavAgent model \cite{zhang2025citynavagent} uses a scoring mechanism based on comparison language-image pre-training (CLIP) and three-dimensional non-maxima suppression (3D-NMS) to simplify subgraphs in large-scale spatial structures, which significantly improves the planning efficiency in such scenarios.

While the above approaches focus on accessible locations in the environment, there is a parallel study that shifts the focus to exploring the elements present in the environment. Instead of just modelling location, this approach breaks down the environment into discrete entities, providing richer information about the agent's interaction with its surroundings. For example, the ConceptGraphs model ~\cite{gu2024conceptgraphs} is built on a 2D foundation model and further expanded into a 3D scene representation, capable of recognizing any object in the scene. By taking each object as a node and giving each node rich semantic properties, the model enables the robot to form a more structured perception of the environment. Similarly, the Open3DSG method \cite{koch2024open3dsg} can directly infer the correlation between objects from the point cloud data without annotating in advance. Finally, the OSU-3DSG model \cite{yu2025osu3dsg} introduces the retrieval-augmented inference process to maintain a scene graph vector database, so that more complex multimodal reasoning and task planning can be completed in a real unstructured environment.

\subsection{Implicit Experiential Memory}

\label{sec: connector-implicit}

The core idea of implicit experiential memory is to integrate non-predefined, semantically organized experience fragments into the spatial framework, so that intelligent systems can reuse the experience in past trajectories \cite{shang2019worlddiscovery,zhu2023vmg}. For example, the WGD model \cite{shang2019worlddiscovery} first identifies the key states in the trajectory and then learns to obtain a world graph that is not related to the specific task. Then, the structural connectivity and traversability of this graph can be used to accelerate the subsequent hierarchical reinforcement learning process. The CSTIG method \cite{verbelen2022cstig} adopts the idea of information geometry for trajectory segmentation. As the sensor collects trajectory data, the model continuously calculates the Fisher information distance to measure the degree of difference between the data. When this distance exceeds the set threshold, a new track fragment (i.e., a new node) is generated. In addition, if the prediction error of the model increases sharply, the trajectory segmentation operation will also be triggered. In this way, a graph can be constructed at the coarse-grained abstraction level of the state without fixed parameters.

In recent years, a number of studies have emerged focusing on extracting more valuable information from these segmented data. The VMG model \cite{zhu2023vmg} transforms offline experience into directed metric chains. It also adds an action translator to transform abstract actions into concrete actions that agents can perform in a real environment. By performing value calculations on this graph, control problems characterized by long-distance and sparse rewards can be transformed into short-distance planning tasks on metric chains. The GBMR method stores the experience in the state-action graph memory module and generates candidate paths by combining trajectory fragments between key nodes, which are used to update value information. Such candidate paths do not introduce new states, but only transfer value between existing nodes. Therefore, even if a specific path is never actually walked, the agent can still navigate along that path. The RLF model \cite{hu2024rlf} uses graph neural networks (GNNs) to learn the long-term changes of the environment from stored memory fragments, resulting in stronger long-term prediction capabilities. Finally, the G4RL model~\cite{zhang2025g4rl} models the spatial topology representation of sub-targets in a graph encoder-decoder architecture. When the agent encounters new or unexplored states, the model can map these states to the learned sub-target space and score them by distance.

\subsection{Discussion}

The core of connector methods is to inject spatial RIB into the topological framework, and then replace the point-by-point search in continuous space with a simpler graph retrieval operation. However, there is a significant flaw in this method: the number of nodes and edges in the graph usually increases almost linearly with the increase of exploration time, which puts great pressure on memory and storage resources. Although some methods \cite{zhang2021l3p,bagatella2022ppgs} can alleviate the expansion of the graph scale, how to deal with the scene of frequent changes in the connection is still an urgent problem to be solved. %How to balance the sparsity of the graph and the integrity of planning and calculation also needs to be further explored. %In addition, these models are sensitive to errors in edge weights. Even small deviations can lead to complete failure of the entire planning process.

\section{Graph as Simulator}

\label{sec: simulator}

\begin{figure}[t]

 \centering

 \includegraphics[width=\linewidth]{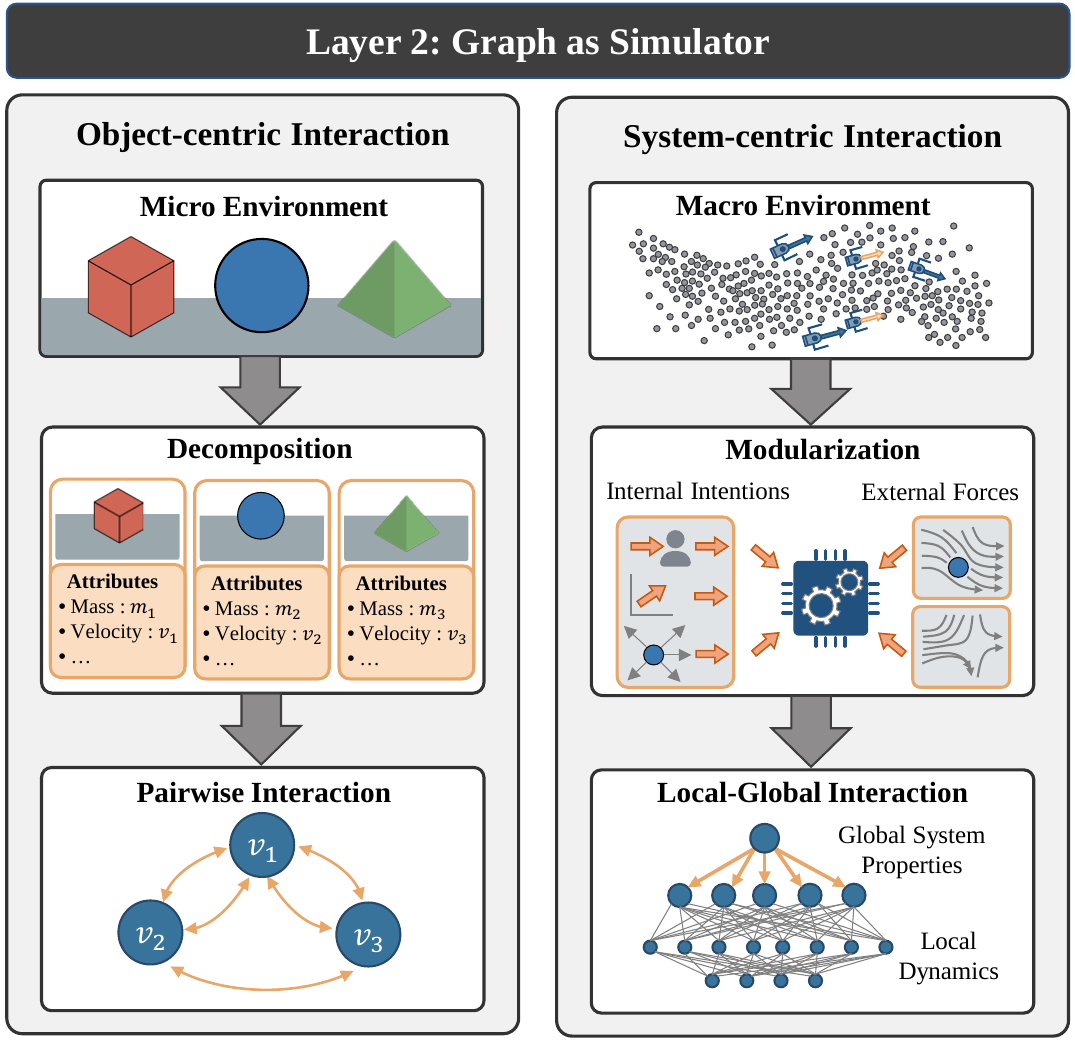}

 \caption{Illustration of graph as simulator. (Left) \textit{Object-centric Interaction}. (Right) \textit{System-centric Interaction}.}

 \label{fig: simulator}

\end{figure}

At the simulator level, methods focus on modeling physical interactions such as collisions and friction to predict how these entities interact over time. The core goal of these methods is to inject physical RIB, which means that they do not try to retain fine motion information at the pixel level, but distill the laws of spatio-temporal variation into a structured interaction model, thereby reducing the risk of error accumulation. When faced with a new scene configuration, the model only needs to apply these interaction rules to the new graph structure to realize the migration of dynamic modes~\cite{kipf2020cswm,whitney2025hdvpd,ugadiarov2025roca}. As shown in \Cref{fig: simulator}, there are two subcategories: \textit{object-oriented interactions} (see \Cref{sec: simulator-object}) and \textit{system-oriented interactions} (see \Cref{sec: simulator-systemic}).

\subsection{Object-centric Interaction}

\label{sec: simulator-object}

The object-centric interaction methods are mainly used to model the physical interaction behavior between a small number of discrete entities. By decomposing observations into independent entities, these methods reduce pixel-level noise. Object-centric models such as C-SWM and G-SWM are foundational examples of simulator-style GWMs, as they explicitly use structured object representations and relational transition functions to model physical dynamics. Specifically, the C-SWM model \cite{kipf2020cswm} uses contrastive learning to identify implicit objects in the environment and models the state transition process with the help of GNNs, eliminating the need to reconstruct pixels throughout the process. In addition, the G-SWM model \cite{lin2020gswm} performs messaging on the object graph to simulate entity interaction and occlusion phenomena, and introduces explicit implicit state variables to capture multimodal uncertainty and environment-specific behaviour patterns. The CWM model \cite{li2020cwm} incorporates an unsupervised denoising mechanism that can infer hidden physical factors and thus support counterfactual predictions. Similarly, the VGPL model \cite{li2020vgpl} introduces an intermediate particle representation layer that maps visual observation data to particle position and grouping information. Under the constraints of prior knowledge from dynamics, the model can not only adapt to unknown physical properties, but also make the long-term prediction results more stable.

Subsequent research focuses on improving the robustness of the simulator and the efficiency of data utilization. In order to reduce the computational overhead, the GATSBI model \cite{min2021gatsbi} clips the graph structure with the help of the $k$ nearest neighbor algorithm. In order to ensure spatial consistency, the 3D-OES model \cite{tung20213does} migrates the relational reasoning process to the 3D voxel space, and the model can still perform the motion prediction task stably even if the perspective changes. The CEE-US model \cite{sancaktar2023ceeus} also introduces an active exploration mechanism, which uses the prediction bias generated by GNNs as intrinsic reward signals to guide agents to explore areas with high uncertainty. In this way, the agent can autonomously discover complex dynamic patterns in the environment, thereby directly supporting zero-shot manipulation tasks.

Recent research progress has focused on finer deconstruction of object properties and model improvements for complex environments. The ROCA model \cite{ugadiarov2025roca} uses pre-trained object representation to construct a graph structure, and embeds the world model directly into the evaluation network to achieve value estimation, which significantly improves the sample utilization efficiency. The FIOC-WM model \cite{feng2025fiocwm} deconstructs the representation at the object and attribute levels, and uses the inverse dynamic model to extract the key features related to interaction behavior. To solve the visual interference problem in complex and chaotic environments, the Dyn-O model \cite{wang2025dyno} distinguishes static and dynamic features, and uses Mamba to improve generalization.

\subsection{System-centric Interaction}

\label{sec: simulator-systemic}

Unlike object-centric approaches, system-centric interaction approaches target complex interconnected environments, such as multi-agent collaboration and large-scale dynamic simulation. In such scenarios, models need to effectively separate the autonomous intention of the agent from the external interference in the environment, and puts forward high requirements for the scalability of modular design.

As an early research in this direction, the VDFD model \cite{wang2023vdfd} designs parallel action condition branches and non-action branches, which can accurately distinguish the agent's autonomous intention and the action of external environmental forces such as water flow in multi-agent collaboration tasks. Recent studies have further expanded the scale of simulation and strengthened the fusion ability of multimodal information. The RoboPack model \cite{ai2024robopack} uses particle graphs to dynamically model robot interaction tasks in real scenarios, and completes physical parameter estimation with historical data from haptic interactions. This method greatly improves the accuracy of long-term prediction and optimizes the performance of the model predictive control system. In the field of dynamic simulation, the HD-VPD model \cite{whitney2025hdvpd} proposes an architecture that stacks the global attention mechanism with the local GNN layer alternately. This design can process more than 100,000 implicit particles in real time, enabling high-fidelity simulation of complex materials.

\subsection{Discussion}

The core of the simulator method is to inject physical RIB into an interaction model. The evolution of simulation technology from object-oriented to system-oriented reflects the increasing research attention to system-level co-evolution, and its research scope covers multiple directions such as multi-agent intention modeling and haptic feedback fusion. However, the current mainstream model architecture is still a deterministic model, which limits its ability to model stochastic dynamic processes in the real physical world.
\begin{figure}[t]

 \centering

 \includegraphics[width=\linewidth]{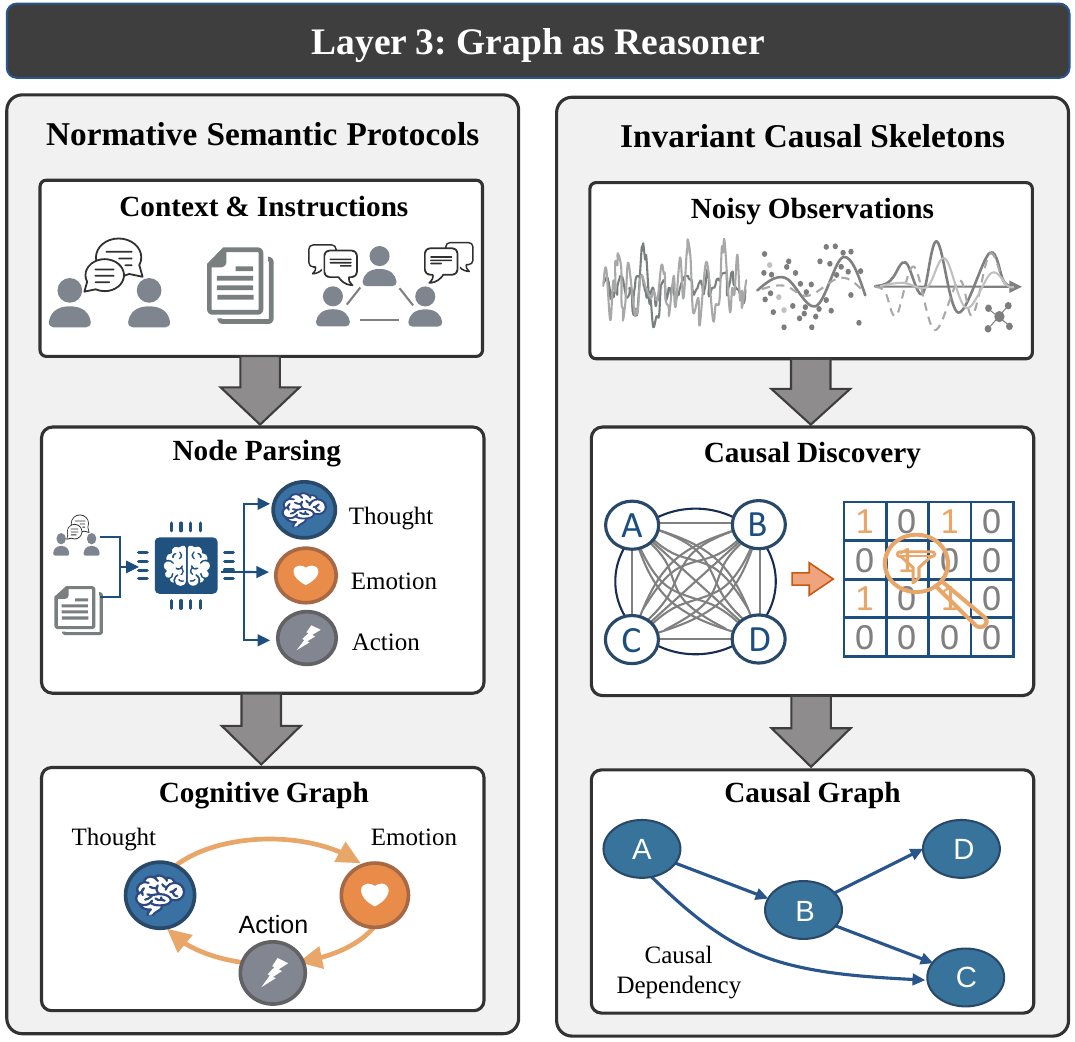}

 \caption{Illustration of graph as reasoner. (Left) \textit{Normative semantic protocols}. (Right) \textit{Invariant causal skeletons}.}

 \label{fig: reasoner}

\end{figure}

\section{Graph as Reasoner}

\label{sec: reasoner}

Although the simulator method takes into account the laws of physics, its reasoning ability is still unstable and lacks an understanding of the mental world~\cite{fung2025embodied}. At the reasoner level, the core goal of the model is to inject logical RIB, that is, to transform the dynamic patterns of the environment into stable logical structures. At this level, nodes $\mathcal{V}$ represent various cognitive concepts or causal factors, while edges $\mathcal{E}$ carry the meaning of semantic constraints or causal associations. As shown in \Cref{fig: reasoner}, the research content in this field can be divided into two main directions: one is the \textit{normative semantic protocols} (see \Cref{sec: reasoner-symbolic}), which defines a structured framework for reasoning. The second is the \textit{invariant causal skeletons} (see \Cref{sec: reasoner-causal}), which is to deduce causality through observational data.

\subsection{Normative Semantic Protocols}\label{sec: reasoner-symbolic}

These methods aim to build structured semantic protocols or cognitive models to support collaboration, task execution, and high-level reasoning among agents. Early works mainly use knowledge graphs or memory graphs to maintain structured internal states in textual and interactive environments. For example, Worldformer~\cite{ammanabrolu2021worldformer} models textual environments with knowledge graphs and predicts graph transitions across states, while AriGraph~\cite{anokhin2025arigraph} integrates semantic and episodic memories into a unified memory graph for LLM-based agents.

Recent studies further extend this idea to social simulation and cognitive modeling, where LLMs are often used as semantic parsers, decision modules, or protocol generators. At the macroscopic level, S3~\cite{gao2023s3} and YuLan-OneSim~\cite{wang2025yulan} construct large-scale agent-oriented social simulators based on social or interaction graphs. At the microscopic level, COKE~\cite{wu2024coke} represents theory-of-mind reasoning as a cognitive graph, Polanyi~\cite{de2025polanyi} incorporates moral and normative knowledge into structured world representations, and SWMPO~\cite{cano2025swmpo} learns symbolic protocols for structured reasoning across different physical modes.

More recently, graph-based semantic protocols have been applied to planning, persistent memory, functional scene understanding, and multimodal world modeling. FPWC~\cite{yin2025fpwc} represents the world as a text-driven directed graph for code-based device control, DAVIS~\cite{dinh2025davis} builds spatio-temporal knowledge graphs as persistent memory for robust decision-making, and OpenFunGraph~\cite{zhang2025openfungraph} constructs open-vocabulary functional 3D scene graphs for physical interaction understanding. Finally, GWM~\cite{feng2025gwm} moves toward multimodal graph foundation models by integrating action nodes, task logic, and environment dynamics through graph-based message passing.

\subsection{Invariant Causal Skeletons}

\label{sec: reasoner-causal}

The invariant causal skeleton is extracted from noisy observation data by hidden space causal discovery technology. The core goal of this approach is to uncover the causal mechanisms that remain stable in different environments or intervention scenarios. This mechanism is often presented as a directed acyclic graph (DAG) or a set of modular causal components. By stripping away false associations and preserving real causal relationships, such models can achieve robust counterfactual reasoning.

For example, the FANS-RL model \cite{feng2022fansrl} models unobserved non-stationary properties in the environment as implicit factors that change over time, and learns the causal structure and representation of such variables under the framework of decomposed Markov decision processes (MDPs). The VCD model combines variational reasoning with differentiable causal discovery technology, which can simultaneously construct binary hidden space causal graphs and identify hidden variables. Under the premise that the soft intervention hypothesis is true, the model can realize the reuse of causal mechanisms in different environments. The FACTS model \cite{nanbo2025facts} is a decomposed state-space world model, which introduces a memory-input routing strategy based on an attention mechanism to correlate and match input features with implicit factors. At the same time, the model is also equipped with a selective historical compression strategy, which enables FACTS to stably track the dynamic changes of various factors. In an open-world environment, the CCSA model \cite{zhao2025ccsa} employs a meta-causal graph to characterize context-dependent causal subgraphs. With the help of curiosity-driven exploration and intervention verification mechanisms, the model can actively explore and calibrate the causal graph, thereby alleviating the problems of insufficient data coverage and ambiguous edge relationships in pure observation scenarios.

\subsection{Discussion}

Although the reasoner methods have strong reasoning ability, existing methods still struggle to balance semantic flexibility and structural rigor. Methods based on semantic protocols have strong interpretability, but the edge relationships generated by LLMs may often not meet geographical or physical constraints, which can lead to logical illusions. Although the autonomous causal discovery method can still maintain stability under external interference, it is difficult to adapt to the complex characteristics of human social scenes. The integration of these two research directions, that is, anchoring semantic nodes on the causal graph corresponding to the verified causal framework, may be an effective way to resolve this contradiction.
\section{Challenges and Future Directions}
% In the previous sections, we have systematically formalized and categorized the emerging field of graph world models, transitioning it from a collection of fragmented studies into a cohesive research paradigm. By introducing a three-layer taxonomy centered on spatial, physical, and logical redundancy compression, we have illustrated how graph structures serve as powerful inductive biases to enhance the predictive accuracy and planning efficiency of autonomous agents. While GWMs have demonstrated superior performance in diverse domains ranging from visual navigation to complex physical reasoning, the field remains in its nascent stages. In this section, we highlight some open challenges and promising future research directions to advance the field.

\subsection{Dynamic and Lifetime Topological Plasticity}

Existing connector approaches tend to build static or slowly expanding topological frameworks \cite{savinov2018sptm,zhang2021l3p}. However, the real-world environment is non-stationary: the navigation path may be temporarily blocked by dynamic obstacles, and the scene may also undergo permanent structural changes \cite{zhang2025citynavagent,zhu2023vmg}. The current GWM lacks a mechanism to forget or correct wrong edges, and once its internal structure does not match reality, the planning is easy to fail. To solve the above problems, future research can explore topological plasticity, so that the graph structure can be adjusted online according to the inconsistency between the perceptual information and the graph. To improve accuracy in the lifelong learning process, it is necessary to transition from a static, distance-driven cropping method to a dynamic graph.

\subsection{Support for Probabilistic Dynamic Modelling}
At present, most simulators use the deterministic state transition method. This means that given the current state and action, only a single future graph structure is predicted. This modelling method cannot adequately characterize high-entropy dynamic processes in the real world such as liquid spills and crowd disorder disturbances~\cite{whitney2025hdvpd}, and may lead to overconfident decision-making behavior in safety-critical areas such as autonomous driving. To address this issue, future research can incorporate graph diffusion models or variational priors into transfer models, enabling GWMs to generate and score multiple plausible future hypotheses. The incorporation of multi-hypothetical inference gives the planning process more credibility while also providing the model with the necessary buffer to deal with the performance degradation that is often caused by distribution shifts in real-world deployments.

\subsection{Mitigating Biases and Logical Hallucinations} 
When LLMs are introduced at the reasoner level, logical hallucinations are prone to occur. Studies related to CityGPT~\cite{feng2025citygpt} and AgentMove~\cite{feng2025agentmove} show that LLMs may generate graph edges that seem semantically reasonable but are geographically invalid (e.g., generate shortcuts that do not exist in reality in complex urban layouts). To solve such problems, future GWMs can add a differentiable verification layer, so that the symbols generated by LLMs are always aligned with the geospatial knowledge graph and physics engine during the inference process. These authenticity constraints allow the inference graph to be anchored on empirical evidence rather than relying on unconstrained text deduction.

\subsection{Injecting Multi-granularity Inductive Biases}
Existing works often struggle to jointly model logical, physical, and low-level spatial interactions~\cite{zhu2023vmg}. However, in the long-term planning task, the model is prone to lose control of the global goal due to the optimization focus on local details. GWMs may benefit from a multilayer architecture. Based on multi-graph learning, future research can design GWMs based on abstract-reality dual-layer graphs, where the abstract layer encodes semantics or causality, and the reality layer models fine-grained waypoints or physical contact behavior. This multi-granularity injection method may help improve long-time domain reasoning and planning capabilities.

\subsection{Benchmarking Graph World Models}

Although some studies have tried to standardize the evaluation system of the world model~\cite{qin2025worldsimbench,huang2025vbench++}, GWM evaluation remains fragmented and overly dependent on task-level success rates, which may obscure structural and relational failures. Future benchmarks should therefore evaluate not only downstream performance but also graph-specific properties. These properties may include graph fidelity for spatial, physical, semantic, or causal relations, as well as relational transition accuracy for node and edge dynamics. They should also consider long-horizon stability, generalization under distribution shift, reasoning correctness, and graph efficiency in construction, update, retrieval, and planning. Such benchmarks would provide more diagnostic protocols for identifying whether failures arise from perception, graph abstraction, relational transition, planning, or reasoning.

\section{Conclusion}

In this survey, we integrate the field from a series of fragmented research efforts into a unified research paradigm by proposing a standardized definition of graph world models (GWMs). This formal definition provides a unified communication context for researchers in the field, which is convenient for subsequent evaluation and expansion of GWMs. Furthermore, based on relational inductive biases, we survey GWMs through a three-layer taxonomy (i.e., connector, simulator, and reasoner) to provide a systematic synthesis of the literature. For each category, we outline the core design principles and conduct an in-depth review, comprehensive analysis, and systematic comparison of existing works. Finally, we highlight open challenges and promising future research directions to advance the field of GWMs.
% \clearpage

% \bibliographystyle{ieeetr.bst}
% \bibliography{main}
% \bibliographystyle{IEEEtran}
\small
\bibliographystyle{IEEEtran}
\bibliography{main}

@article{ding2025understanding,
  title={Understanding world or predicting future? a comprehensive survey of world models},
  author={Ding, Jingtao and Zhang, Yunke and others},
  journal={CSUR},
  volume={58},
  number={3},
  pages={1--38},
  year={2025},
  publisher={ACM New York, NY}
}

@article{ha2018world,
  title={World Models},
  author={Ha, David and Schmidhuber, J{\"u}rgen},
  journal={arXiv preprint arXiv:1803.10122},
  year={2018}
}

@inproceedings{savinov2018sptm,
  title={Semi-parametric topological memory for navigation},
  author={Savinov, Nikolay and Dosovitskiy, Alexey and Koltun, Vladlen},
  booktitle={Proc. of ICLR},
  year={2018}
}

@inproceedings{eysenbach2019sorb,
  title={Search on the replay buffer: bridging planning and reinforcement learning},
  author={Eysenbach, Benjamin and Salakhutdinov, Ruslan and Levine, Sergey},
  booktitle={Proc. of NeurIPS},
  pages={15246--15257},
  year={2019}
}

@inproceedings{taniguchi2021point,
  title={Pose invariant topological memory for visual navigation},
  author={Taniguchi, Asuto and Sasaki, Fumihiro and Yamashina, Ryota},
  booktitle={Proc. of ICCV},
  pages={15384--15393},
  year={2021}
}

@inproceedings{bagatella2022ppgs,
  title={Planning from Pixels in Environments with Combinatorially Hard Search Spaces},
  author={Bagatella, Marco and Ol{\v{s}}{\'a}k, Mirek and Rolinek, Michal and Martius, Georg},
  booktitle={Proc. of NeurIPS},
  pages={24707--24718},
  year={2022}
}

@inproceedings{verbelen2022cstig,
  title={Chunking space and time with information geometry},
  author={Verbelen, Tim and De Tinguy, Daria and others},
  booktitle={NeurIPS 2022 Workshop InfoCog},
  year={2022}
}

@inproceedings{wang2023dreamwalker,
  title={Dreamwalker: Mental planning for continuous vision-language navigation},
  author={Wang, Hanqing and Liang, Wei and Van Gool, Luc and Wang, Wenguan},
  booktitle={Proc. of ICCV},
  pages={10873--10883},
  year={2023}
}

@inproceedings{zhao2024overnav,
  title={OVER-NAV: Elevating Iterative Vision-and-Language Navigation with Open-Vocabulary Detection and StructurEd Representation},
  author={Zhao, Ganlong and Li, Guanbin and others},
  booktitle={Proc. of CVPR},
  pages={16296--16306},
  year={2024}
}

@inproceedings{zhang2025citynavagent,
  title={CityNavAgent: Aerial Vision-and-Language Navigation with Hierarchical Semantic Planning and Global Memory},
  author={Zhang, Weichen and Gao, Chen and Yu, Shiquan and others},
  booktitle = {Proc. of ACL},
  year={2025}
}

@article{shang2019worlddiscovery,
  title={Learning world graphs to accelerate hierarchical reinforcement learning},
  author={Shang, Wenling and Trott, Alex and Zheng, Stephan and others},
  journal={arXiv preprint arXiv:1907.00664},
  year={2019}
}

@inproceedings{zhang2021l3p,
  title={World model as a graph: Learning latent landmarks for planning},
  author={Zhang, Lunjun and Yang, Ge and Stadie, Bradly C},
  booktitle={Proc. of ICML},
  pages={12611--12620},
  year={2021},
  organization={PMLR}
}

@article{kang2023gbmr,
  title={Sample efficient reinforcement learning using graph-based memory reconstruction},
  author={Kang, Yongxin and Zhao, Enmin and Zang, Yifan and others},
  journal={IEEE Trans. Artif. Intell},
  volume={5},
  number={2},
  pages={751--762},
  year={2023},
  publisher={IEEE}
}

@inproceedings{bonnavaud2023rgl,
  title={Learning State Reachability as a Graph in Translation Invariant Goal-based Reinforcement Learning Tasks},
  author={Bonnavaud, Hedwin and Albore, Alexandre and Rachelson, Emmanuel},
  booktitle={EWRL},
  year={2023}
}

@inproceedings{zhu2023vmg,
  title={Value Memory Graph: A Graph-Structured World Model for Offline Reinforcement Learning},
  author={Zhu, Deyao and Li, Li Erran and others},
  booktitle={Proc. of ICLR},
  year={2023}
}

@article{zhang2025g4rl,
  title={Incorporating Spatial Information into Goal-Conditioned Hierarchical Reinforcement Learning via Graph Representations},
  author={Zhang, Shuyuan and Wang, Zihan and Chang, Xiao-Wen and Precup, Doina},
  journal={TMLR},
  year={2025}
}

@inproceedings{tung20213does,
  title={3D-OES: Viewpoint-Invariant Object-Factorized Environment Simulators},
  author={Tung, Hsiao-Yu and Xian, Zhou and others},
  booktitle={Proc. of CoRL},
  pages={1669--1683},
  year={2021},
  organization={PMLR}
}

@inproceedings{kipf2020cswm,
  title={Contrastive Learning of Structured World Models},
  author={Kipf, Thomas and van der Pol, Elise and Welling, Max},
  booktitle={Proc. of ICLR},
  year={2020}
}

@inproceedings{lin2020gswm,
  title={Improving generative imagination in object-centric world models},
  author={Lin, Zhixuan and Wu, Yi-Fu and others},
  booktitle={Proc. of ICML},
  pages={6140--6149},
  year={2020},
  organization={PMLR}
}

@inproceedings{li2020vgpl,
  title={Visual grounding of learned physical models},
  author={Li, Yunzhu and Lin, Toru and Yi, Kexin and others},
  booktitle={Proc. of ICML},
  pages={5927--5936},
  year={2020},
  organization={PMLR}
}

@inproceedings{min2021gatsbi,
  title={Gatsbi: Generative agent-centric spatio-temporal object interaction},
  author={Min, Cheol-Hui and Bae, Jinseok and others},
  booktitle={Proc. of CVPR},
  pages={3074--3083},
  year={2021}
}

@inproceedings{sancaktar2023ceeus,
  title={Curious Exploration via Structured World Models Yields Zero-Shot Object Manipulation},
  author={Sancaktar, Cansu and Blaes, Sebastian and Martius, Georg},
  booktitle={Proc. of NeurIPS},
  pages={24170--24183},
  year={2022}
}

@article{hu2024rlf,
  title={Invariant Representations Learning with Future Dynamics},
  author={Hu, Wenning and He, Ming and Chen, Xirui and Wang, Nianbin},
  journal={Eng. Appl. Artif. Intell},
  volume={128},
  pages={107496},
  year={2024},
  publisher={Elsevier}
}

@article{fung2025embodied,
  title={Embodied ai agents: Modeling the world},
  author={Fung, Pascale and Bachrach, Yoram and Celikyilmaz, Asli and others},
  journal={arXiv preprint arXiv:2506.22355},
  year={2025}
}

@inproceedings{ai2024robopack,
  title={RoboPack: Learning Tactile-Informed Dynamics Models for Dense Packing},
  author={Ai, Bo and Tian, Stephen and Shi, Haochen and others},
  booktitle={ICRA 2024 Workshop on 3D Visual Representations for Robot Manipulation},
  year={2024}
}

@inproceedings{whitney2025hdvpd,
  title={Modeling the Real World with High-Density Visual Particle Dynamics},
  author={Whitney, William F and Varley, Jake and others},
  booktitle={Proc. of CoRL},
  pages={1427--1442},
  year={2024}
}

@inproceedings{ugadiarov2025roca,
  title={Relational object-centric actor-critic},
  author={Ugadiarov, Leonid Anatolievich and Vorobyov, Vitaliy and Panov, Aleksandr},
  booktitle={Proc. of CLeaR},
  year={2025}
}

@inproceedings{nanbo2025facts,
  title={FACTS: A Factored State-Space Framework for World Modelling},
  author={Nanbo, Li and Laakom, Firas and Wang, Wenyi and Schmidhuber, J{\"u}rgen and others},
  booktitle={Proc. of ICLR},
  year={2025}
}

@inproceedings{feng2025fiocwm,
  title={Learning Interactive World Model for Object-Centric Reinforcement Learning},
  author={Feng, Fan and Lippe, Phillip and Magliacane, Sara},
  booktitle={Proc. of NeurIPS},
  year={2025}
}

@article{wang2023vdfd,
  title={Leveraging world model disentanglement in value-based multi-agent reinforcement learning},
  author={Wang, Zhizun and Meger, David},
  journal={arXiv preprint arXiv:2309.04615},
  year={2023}
}

@article{li2020cwm,
  title={Causal world models by unsupervised deconfounding of physical dynamics},
  author={Li, Minne and Yang, Mengyue and Liu, Furui and Chen, Xu and others},
  journal={arXiv preprint arXiv:2012.14228},
  year={2020}
}

@article{lei2023vcd,
  title={Variational Causal Dynamics: Discovering Modular World Models from Interventions},
  author={Lei, Anson and Sch{\"o}lkopf, Bernhard and Posner, Ingmar},
  journal={TMLR},
  year={2023}
}

@article{zhao2025ccsa,
  title={Curious Causality-Seeking Agents Learn Meta Causal World},
  author={Zhao, Zhiyu and Li, Haoxuan and Zhang, Haifeng and others},
  journal={arXiv preprint arXiv:2506.23068},
  year={2025}
}

@inproceedings{ammanabrolu2021worldformer,
  title={Learning knowledge graph-based world models of textual environments},
  author={Ammanabrolu, Prithviraj and Riedl, Mark O},
  booktitle={Proc. of NeurIPS},
  pages={3720--3731},
  year={2021}
}

@inproceedings{feng2022fansrl,
  title={Factored adaptation for non-stationary reinforcement learning},
  author={Feng, Fan and Huang, Biwei and Zhang, Kun and Magliacane, Sara},
  booktitle={Proc. of NeurIPS},
  pages={31957--31971},
  year={2022}
}

@article{gao2023s3,
  title={S3: Social-network simulation system with large language model-empowered agents},
  author={Gao, Chen and Lan, Xiaochong and Lu, Zhihong and Mao, Jinzhu and others},
  journal={arXiv preprint arXiv:2307.14984},
  year={2023}
}

@inproceedings{wu2024coke,
  title={Coke: A cognitive knowledge graph for machine theory of mind},
  author={Wu, Jincenzi and Chen, Zhuang and Deng, Jiawen and Sabour, Sahand and Meng, Helen and Huang, Minlie},
  booktitle={Proc. of ACL},
  pages={15984--16007},
  year={2024}
}

@inproceedings{fu2024anyhome,
  title={Anyhome: Open-vocabulary generation of structured and textured 3d homes},
  author={Fu, Rao and Wen, Zehao and Liu, Zichen and others},
  booktitle={Proc. of ECCV},
  pages={52--70},
  year={2024}
}

@inproceedings{feng2025citygpt,
  title={Citygpt: Empowering urban spatial cognition of large language models},
  author={Feng, Jie and Liu, Tianhui and Du, Yuwei and others},
  booktitle={Proc. of KDD},
  pages={591--602},
  year={2025}
}

@inproceedings{feng2025agentmove,
  title={Agentmove: A large language model based agentic framework for zero-shot next location prediction},
  author={Feng, Jie and Du, Yuwei and Zhao, Jie and Li, Yong},
  booktitle={Proc. of NAACL},
  pages={1322--1338},
  year={2025}
}

@article{de2025polanyi,
  title={Neurosymbolic graph enrichment for grounded world models},
  author={De Giorgis, Stefano and Gangemi, Aldo and others},
  journal={IPM},
  volume={62},
  number={4},
  pages={104127},
  year={2025},
  publisher={Elsevier}
}

@inproceedings{dinh2025davis,
  title={DAVIS: Planning Agent with Knowledge Graph-Powered Inner Monologue},
  author={Dinh, Minh Pham and Yankoski, Michael G and Syed, Munira and Ford, Trenton W},
  booktitle={Findings of EMNLP},
  pages={16490--16505},
  year={2025}
}

@article{yin2025fpwc,
  title={Unlocking Smarter Device Control: Foresighted Planning with a World Model-Driven Code Execution Approach},
  author={Yin, Xiaoran and Luo, Xu and Wu, Hao and Gao, Lianli and Song, Jingkuan},
  journal={arXiv preprint arXiv:2505.16422},
  year={2025}
}

@inproceedings{feng2025gwm,
  title={Graph World Model},
  author={Feng, Tao and Wu, Yexin and Lin, Guanyu and You, Jiaxuan},
  booktitle={Proc. of ICML},
  year={2025}
}

@inproceedings{wang2025yulan,
  title={Yulan-onesim: Towards the next generation of social simulator with large language models},
  author={Wang, Lei and Gao, Heyang and Bo, Xiaohe and others},
  booktitle={NeurIPS 2025 Workshop SEA},
  year={2025}
}

@article{guan2024world,
  title={World models for autonomous driving: An initial survey},
  author={Guan, Yanchen and Liao, Haicheng and Li, Zhenning and others},
  journal={IEEE Trans Intell Veh},
  year={2024},
  publisher={IEEE}
}

@article{cho2024sora,
  title={Sora as an agi world model? a complete survey on text-to-video generation},
  author={Cho, Joseph and Puspitasari, Fachrina Dewi and Zheng, Sheng and others},
  journal={arXiv preprint arXiv:2403.05131},
  year={2024}
}

@article{sakagami2023robotic,
  title={Robotic world models—conceptualization, review, and engineering best practices},
  author={Sakagami, Ryo and Lay, Florian S and D{\"o}mel, Andreas and others},
  journal={Front. robot. AI},
  volume={10},
  pages={1253049},
  year={2023},
  publisher={Frontiers Media SA}
}

@article{ha2018recurrent,
  title={Recurrent world models facilitate policy evolution},
  author={Ha, David and Schmidhuber, J{\"u}rgen},
  journal={Proc. of NeurIPS},
  volume={31},
  year={2018}
}

@inproceedings{qin2025worldsimbench,
  title={WorldSimBench: Towards Video Generation Models as World Simulators},
  author={Qin, Yiran and Shi, Zhelun and Yu, Jiwen and others},
  booktitle={Proc. of ICML},
  year={2025}
}

@misc{OpenAI2024Sora,
author = {OpenAI},
year = {2024},
title = {Sora: Creating video from text},
howpublished = {\url{https://openai.com/sora}},
note = {Retrieved May 06, 2025}
}

@article{hafner2025mastering,
  title={Mastering diverse control tasks through world models},
  author={Hafner, Danijar and Pasukonis, Jurgis and Ba, Jimmy and Lillicrap, Timothy},
  journal={Nature},
  pages={1--7},
  year={2025}
}

@article{battaglia2018relational,
  title={Relational inductive biases, deep learning, and graph networks},
  author={Battaglia, Peter W and Hamrick, Jessica B and Bapst, Victor and others},
  journal={arXiv preprint arXiv:1806.01261},
  year={2018}
}

@article{huang2025vbench++,
  title={Vbench++: Comprehensive and versatile benchmark suite for video generative models},
  author={Huang, Ziqi and Zhang, Fan and Xu, Xiaojie and others},
  journal={IEEE TPAMI},
  year={2025},
  publisher={IEEE}
}

@inproceedings{koch2024open3dsg,
  title={Open3dsg: Open-vocabulary 3d scene graphs from point clouds with queryable objects and open-set relationships},
  author={Koch, Sebastian and Vaskevicius, Narunas and others},
  booktitle={Proc. of CVPR},
  pages={14183--14193},
  year={2024}
}

@inproceedings{gu2024conceptgraphs,
  title={Conceptgraphs: Open-vocabulary 3d scene graphs for perception and planning},
  author={Gu, Qiao and Kuwajerwala, Ali and others},
  booktitle={Proc. of ICRA},
  pages={5021--5028},
  year={2024},
}

@inproceedings{zhang2025openfungraph,
  title={Open-vocabulary functional 3d scene graphs for real-world indoor spaces},
  author={Zhang, Chenyangguang and Delitzas, Alexandros and Wang, Fangjinhua and others},
  booktitle={Proc. of CVPR},
  pages={19401--19413},
  year={2025}
}

@inproceedings{anokhin2025arigraph,
  title={Arigraph: Learning knowledge graph world models with episodic memory for llm agents},
  author={Anokhin, Petr and Semenov, Nikita and Sorokin, Artyom and others},
  booktitle={Proc. of IJCAI},
  year={2025}
}

@inproceedings{wang2025dyno,
  title={Dyn-O: Building Structured World Models with Object-Centric Representations},
  author={Wang, Zizhao and Wang, Kaixin and Zhao, Li and others},
  booktitle={Proc. of NeurIPS},
  year={2025}
}

@inproceedings{yu2025osu3dsg,
  title={Open-World 3D Scene Graph Generation for Retrieval-Augmented Reasoning},
  author={Yu, Fei and Deng, Quan and Tang, Shengeng and others},
  booktitle={Proc. of AAAI},
  year={2026}
}

@inproceedings{cano2025swmpo,
  title={Neurosymbolic World Models for Sequential Decision Making},
  author={Cano, Leonardo Hernandez and Perroni-Scharf, Maxine and Dhir, Neil and others},
  booktitle={Proc. of KDD},
  year={2025}
}

@article{chu2026awm,
  title={Agentic World Modeling: Foundations, Capabilities, Laws, and Beyond},
  author={Chu, Meng and Zhang, Xuan Billy and Lin, Kevin Qinghong and Kong, Lingdong and Zhang, Jize and Tu, Teng and Ma, Weijian and Huang, Ziqi and Yang, Senqiao and Huang, Wei and others},
  journal={arXiv preprint arXiv:2604.22748},
  year={2026}
}

\end{document}